\documentclass[lettersize,journal]{IEEEtran}
\usepackage{amsmath,amsfonts}
\usepackage{algorithmic}
\usepackage{array}
\usepackage[caption=false,font=normalsize,labelfont=sf,textfont=sf]{subfig}
\usepackage{textcomp}
\usepackage{stfloats}
\usepackage{url}
\usepackage{verbatim}
\usepackage{graphicx}
\usepackage{booktabs} 

\usepackage{comment}
\usepackage{amssymb}
\usepackage{multirow}
\usepackage{multicol}

\DeclareMathOperator*{\argmin}{arg\,min}

\usepackage[pagebackref,breaklinks,colorlinks]{hyperref}

\def\BibTeX{{\rm B\kern-.05em{\sc i\kern-.025em b}\kern-.08em
    T\kern-.1667em\lower.7ex\hbox{E}\kern-.125emX}}
\usepackage{balance}
\begin{document}
\title{Effective End-to-End Vision Language Pretraining with Semantic Visual Loss}

\author{Xiaofeng Yang, Fayao Liu, Guosheng Lin 
\thanks{Corresponding author: Guosheng Lin.}
\thanks{Xiaofeng Yang and Guosheng Lin are with School of Computer Science and Engineering, Nanyang Technological University (NTU), Singapore 639798 (email: xiaofeng001@e.ntu.edu.sg, gslin@ntu.edu.sg)}

\thanks{Fayao Liu is with Agency for Science, Technology and Research (A*STAR), Singapore 138632 (email: fayaoliu@gmail.com)}
}

\markboth{Journal of \LaTeX\ Class Files,~Vol.~18, No.~9, September~2020}%
{How to Use the IEEEtran \LaTeX \ Templates}

\maketitle
\begin{abstract}
 Current vision language pretraining models are dominated by methods using region visual features extracted from object detectors. Given their good performance, the extract-then-process pipeline significantly restricts the inference speed and therefore limits their real-world use cases. However, training vision language models from raw image pixels is difficult, as the raw image pixels give much less prior knowledge than region features. In this paper, we systematically study how to leverage auxiliary visual pretraining tasks to help training end-to-end vision language models. We introduce three types of visual losses that enable much faster convergence and better finetuning accuracy. Compared with region feature models, our end-to-end models could achieve similar or better performance on down-stream tasks and run more than 10 times faster during inference. Compared with other end-to-end models, our proposed method could achieve similar or better performance when pretrained for only 10\% of the pretraining GPU hours.
\end{abstract}

\section{Introduction}

Using region visual features~\cite{anderson2018bottom} in vision language tasks is one of the milestones in developing vision language models. The training process of object detectors brings rich object classes and attributes information to the region features. The use of region features significantly reduces the learning difficulty and improves the performance of various tasks, and it has now become a common practice in VQA~\cite{anderson2018bottom,hudson2019gqa}, visual captioning~\cite{anderson2018bottom,luo2018discriminability}, and vision language pretraining~\cite{lu2019vilbert,chen2019uniter}. Given its promising performance, using region visual features has several drawbacks. First, extracting region visual features with an object detector is time-consuming, leading to the difficulty of deploying vision language systems in real-world use-cases. Second, developing an effective feature extraction backbone is tricky and requires careful human engineering. 
\begin{figure}[]
\centering
\includegraphics[width=0.5\textwidth,trim=0 0 0 60,clip]{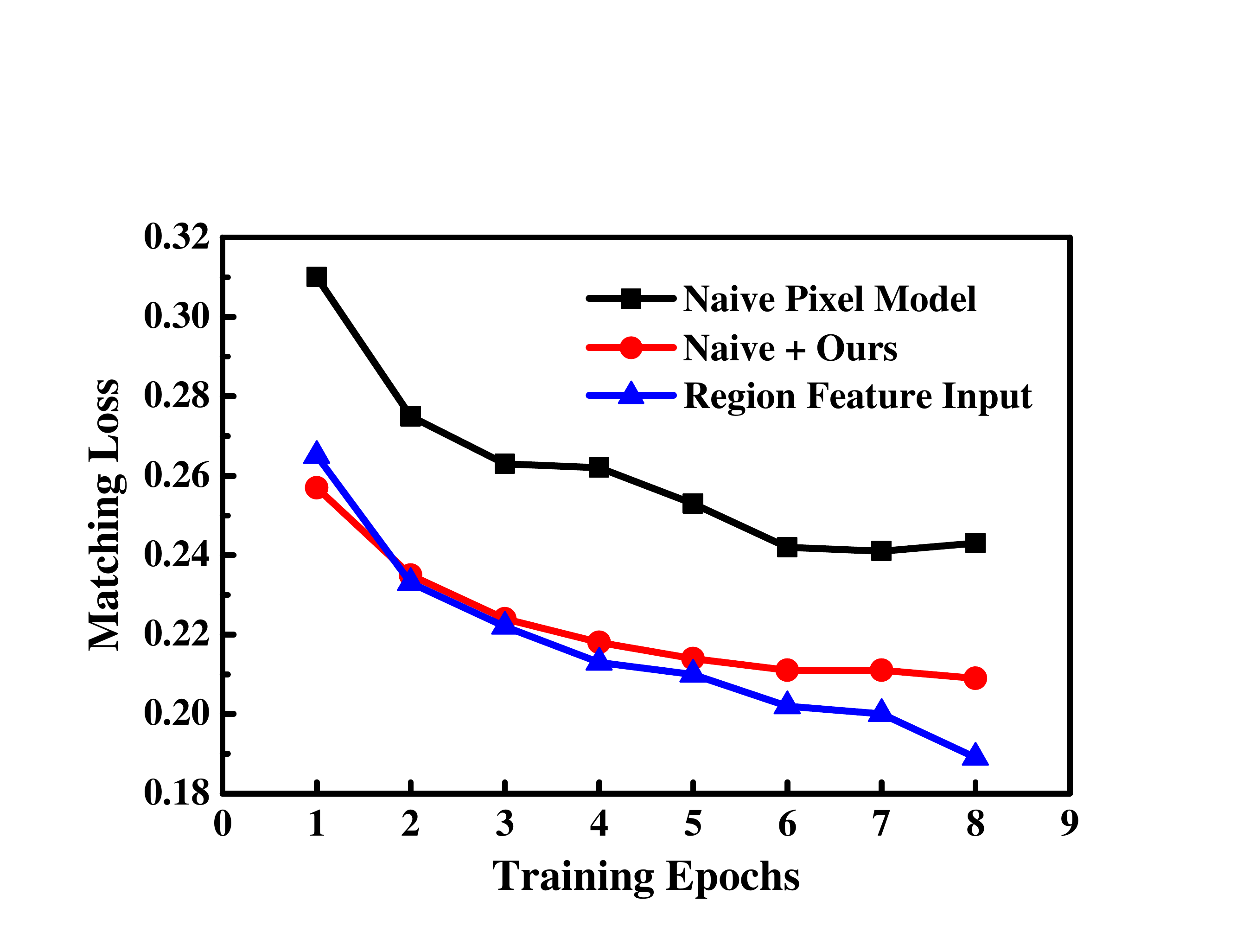}
\caption{Comparison of validation loss during pretraining. A naive pixel model is trained by simply replacing the faster rcnn with a grid feature extractor. Results show that the naive pixel model converges slow and has a high final loss. Once our proposed method is used, the pixel input model could achieve a similar convergence speed as a region feature model. }
\label{fig:1}
\end{figure}

However, training vision language models from raw image pixels is difficult. For existing region feature based methods, region features are extracted from a trained object detector and they contain high-level semantic information and object instance information. In contrast, when we only use raw pixels as input, the high level semantic information and object instance information are missing. In experiments, we demonstrate that simply replacing the faster-rcnn feature extractor~\cite{anderson2018bottom} with a grid feature extraction backbone has the problems of slow convergence and low finetuning accuracy. We attribute this observation to the lack of semantic guidance of pixel inputs during pretraining, i.g. telling the model that the image region is semantically a ``bird''. To this end, we introduce three vision-wise supervision methods to help the training of vision-language models. When used in pretraining, all three losses enable faster convergence and lead to better performance. We show a comparison of validation matching loss during pretraining in Fig~\ref{fig:1}.

First, we propose to use the \textbf{Self-Supervised Loss (SSUL)}. The SSUL trains the vision branch by randomly masking image regions and predicting the mean color (average RGB values) of the masked area. The Self-Supervised Loss does not require additional annotations, hence can be used on any form of image data. Second, we propose to use the \textbf{Semantic Segmentation Loss (SEGL)}. Specifically, we use an extra semantic segmentation loss to supervise the training of the visual branch. The advantage of this loss is that it gives pixel-level semantic supervision to the visual branch. We use the annotation in COCO Stuff~\cite{caesar2018cvpr} to calculate this loss. With only half of the vision language pretraining images annotated, we see a great improvement in convergence speed and finetuning accuracy. Finally, we propose the \textbf{Knowledge Distillation Based Set Prediction Loss (SPL)}. Knowledge distillation~\cite{hinton2015distilling} is an important method to extract knowledge from large models and guide the training of small models. For SPL, we first run the pretrained faster-rcnn model~\cite{anderson2018bottom} to extract object labels from the images. We regard the generated labels as a set of pseudo-labels. The SPL first processes the generated grid feature map with a transformer model to generate a fixed set of region predictions. Then SPL learns to match the generated region predictions with pseudo-labels via optimal matching. The SPL does not require additional human annotations on pretraining images. 

For experiments, we show that our proposed methods could achieve similar performance compared with the region feature models. During inference time, our model runs more than 10 times faster than the region feature models. Compared with other pixel input vision language pretraining models~\cite{huang2020pixel}, our models only require 1/10 of the pre-training GPU hours and get better finetuning accuracy. We also evaluate different vision backbones, e.g. the CNN backbones~\cite{he2016deep} and Vision Transformer backbones~\cite{dosovitskiy2020image}.  

To summarize our contributions:

\begin{itemize}
    \item We study the difficulty in training pixel input vision language pretraining models and propose three types of visual losses to tackle the training difficulty and improve pretraining efficiency. 
    \item Compared with region feature based models, our method achieves similar or better performance and inferences 10 times faster.
    \item Compared with other raw pixel input models, our method trains 10 times faster and achieves similar or better performance.
\end{itemize}

\section{Related Work}

\subsection{Vision Language Pretraining Models}

Unlike traditional vision language methods~\cite{anderson2018bottom,zhang2018high,yan2019stat,hu2020heterogeneous,yuan2020adversarial,yu2020reasoning} that learn vision language mapping based on specific tasks, vision language pretraining models learn robust cross-modal mapping through large scale pretraining.
Vision language pretraining models can be categorized into two types: the one-stream models and two-stream models. The one-stream models~\cite{chen2019uniter,li2020unicoder} first embed the languages and images into embeddings. The language embeddings and image embeddings are then concatenated and processed with a shared transformer network. Since the model parameters are shared, the one-stream models usually have a small model size. However, since the self-attention models like transforms have quadratic time complexity, the one-stream models will have slightly larger memory usage. The two-stream models~\cite{tan2019lxmert,lu2019vilbert,lu202012} process language embeddings and image embeddings with separate transformer encoders and then combine the two features with cross-attention transformer. Compared with the one-stream models, they usually have a larger model size. 

In this paper, we briefly follow the network structure of two-stream models~\cite{tan2019lxmert,lu2019vilbert,lu202012}. We change the region feature input to pixel input and change the visual feature encoder to feature extraction backbones.

\subsection{Loss in Vision Language Pretraining}
Effective loss functions lie in the heart of visual language pretraining. A region feature based vision language model usually involves three types of losses: the language loss, the visual loss, and the matching loss. Most of the time~\cite{li2020unicoder,chen2019uniter,tan2019lxmert,lu2019vilbert}, the language loss is similar to the language modeling loss as in BERT~\cite{devlin2018bert}, in which the language tokens are masked randomly and the models are trained to predict the masked tokens based on two-directional contexts. The visual loss is to mimic the language loss and is done by masking the visual features and predicting the masked regions' classes or attributes. The matching loss predicts if the language input and the vision input are matched. 

When training raw image input based vision language models, the visual inputs (image pixels) don't have specific class annotations. So existing works~\cite{huang2020pixel} only use the language loss and the matching loss. In this paper, we show that such training regimes lead to a slow convergence speed and low finetuning performance. With our proposed visual losses, the raw input models could achieve as fast convergence speed as region feature models in early training stages.
\begin{figure*}[t]
\begin{center}
\includegraphics[width=1\linewidth]{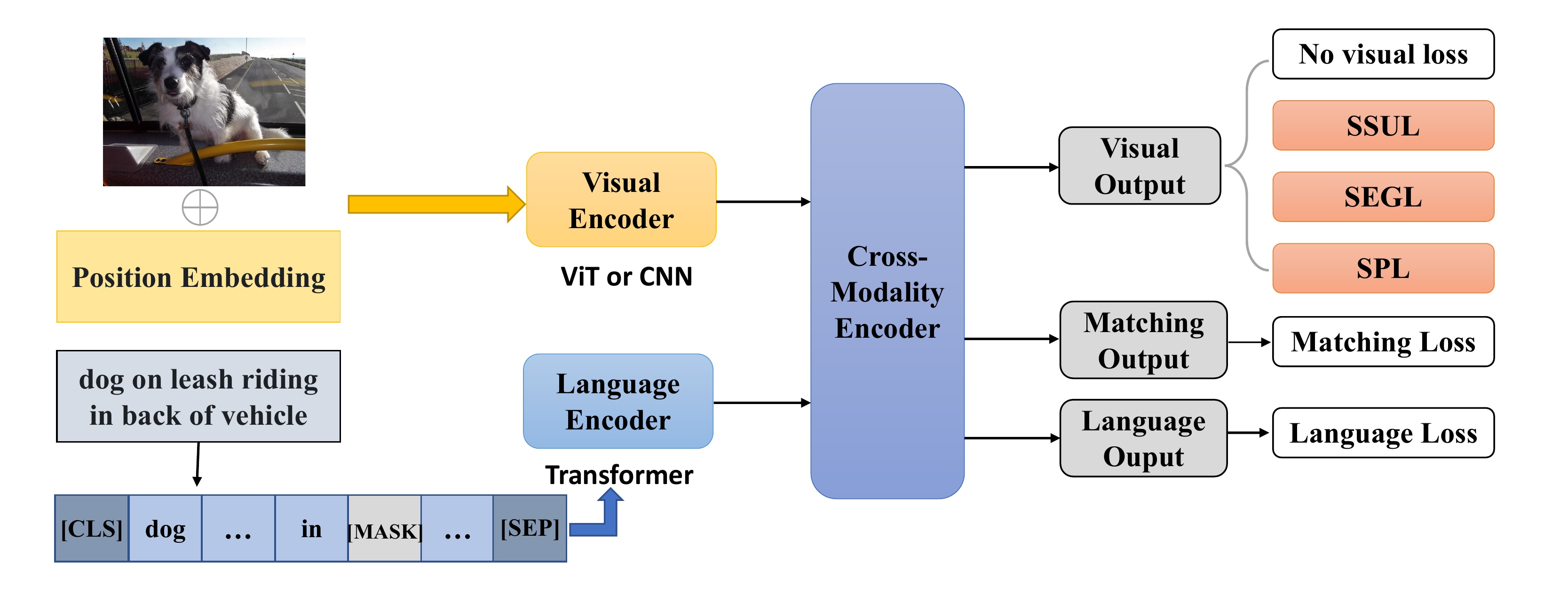}
	\caption{The general network structure. The network takes a raw pixel image and a sentence as input. The image will pass through a visual encoder to extract features. The language will be tokenized and add a [CLS] and a [SEP] tokens. Word tokens are randomly masked with [MASK] tokens for masked language modeling loss. The visual and language features are merged with a cross-modality encoder. Similar to other methods, we use a matching loss and language modeling loss to train the network. Unlike other pixel input vision language models that use no visual loss, we apply our novel visual losses on the visual output.}
	\label{fig:2}
\end{center}
\end{figure*}

\subsection{Efficient Training of Transformer Models}
Transformers~\cite{vaswani2017attention}, as the most widely used language models and vision language models, although perform well in terms of accuracy, take a long training time. Training and testing transformer models efficiently is a long-standing goal in both NLP and computer vision. Existing works can be roughly divided into three categories. The first category~\cite{kitaev2020reformer,zoph2020rethinking,wang2020linformer} 
focuses on improving network structures. They try to optimize the quadratic time complexity of the original transformer model to linear time complexity. These works directly reduce time cost on both training and testing time. The second category~\cite{sanh2019distilbert,jiao2019tinybert,sun2019patient} attempts to extract effective sub-networks from a trained transformer model through knowledge distillations. The third category~\cite{liu2019roberta,gunel2020supervised} focuses on improving the training strategy or proposing new losses. This type of method primarily improves efficiency during training time. 

This paper falls in the third category. We are not trying to propose new network structures or extract sub-networks. We improve the training effectiveness by introducing three visual losses in vision language pretraining.

\section{Effective Visual Loss for Pixel Input Vision Language Pretraining}
In this section, we describe the network structure and the proposed three visual pretraining losses, namely Self-Supervised Loss (SSUL), Semantic Segmentation Loss (SEGL), and Knowledge Distillation Based Set Prediction Loss (SPL). 

\subsection{Network Structure and Other Training Losses}
The general network architecture is shown in Figure~\ref{fig:2}. We briefly follow the network structure of two-stream models.

\begin{figure}[]
\centering
\includegraphics[width=0.5\textwidth,trim=0 0 0 0,clip]{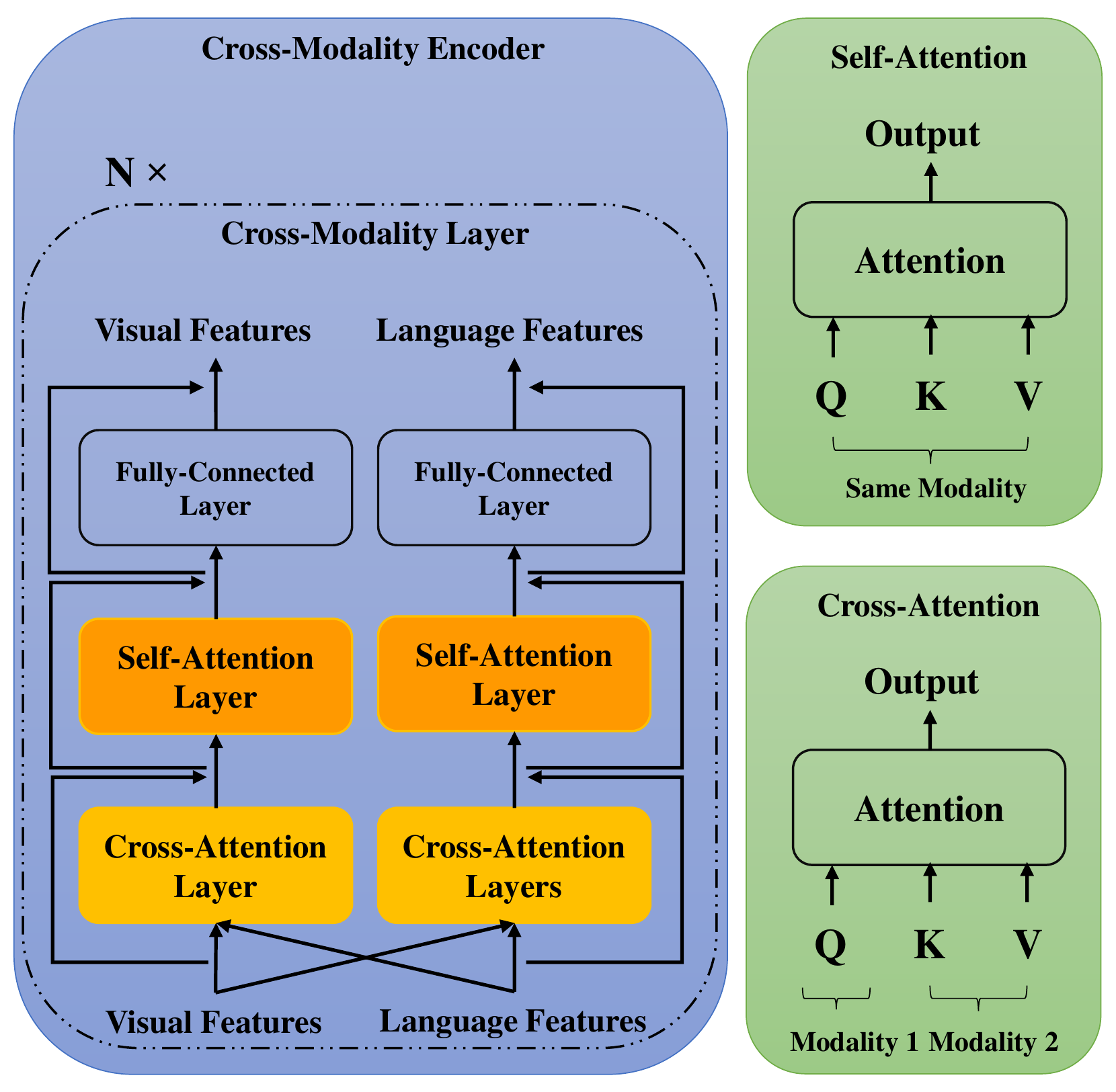}
\caption{Illustration of Cross-Modality Encoder structure. The cross-modality encoder is composed of a sequence of cross-modality layers. We show the detailed structure of the cross-modality layer and show the difference between self-attention and cross-attention mechanisms.}
\label{fig:cross}
\end{figure}

\textbf{Network Structure.}
The image and text inputs are first processed by two separate encoders to extract high level features. The visual features and language features are then passed to the cross-modality encoder. An illustration of the cross-modality encoder structure and the cross-attention mechanism can be found in Figure~\ref{fig:cross}. The cross-modality encoder is composed of cross-modality layers. In each cross-modality layer, the language features and image features are first merged by a cross-attention sub-layer. The cross-attention sub-layer has the same structure as a self-attention layer but uses different inputs as the query and key vectors. For the language branch, the cross-attention sub-layer uses language features from the previous layer as the query vector and image features as the key vector and value vector. After the cross-attention sub-layer, the features are further processed by a self-attention sub-layer and a fully-connected layer. We add short-cut connections between the sub-layers.

The visual input of the network is a raw pixel image. For vision transformer backbone (ViT)~\cite{dosovitskiy2020image}, we slice the image into patches and add an extra learnable position embedding. For CNN backbone, we first run the feature extraction then combine the extracted features with position embedding. 

During pretraining, we follow the common practice to use a language modeling loss and a vision language matching loss.

\textbf{Masked Language Modeling (MLM).} 
Given an input sentence, the words are randomly masked at a ratio of 0.15. Following BERT~\cite{devlin2018bert}, at 80\% of chance, the masked token is changed to [MASK] token. Otherwise, the token is changed to zero or kept as the original token. The masked language modeling loss predicts the masked tokens given bidirectional contexts.

 \textbf{Image-Text Matching.} 
Given the input sentence and image, the image-text matching loss predicts whether the text could describe the image i.g. if the image and text are matched. Fake sentences and real sentences are sampled at 50\%. We use the language feature at [CLS] token to perform the matching task.

For visual loss, existing pixel input vision language pretraining models~\cite{huang2020pixel} can be seen as ``No Visual Loss''. We describe our proposed three losses in the subsection below. Also, we draw the detailed network structure for each loss in Fig~\ref{fig:compare_methods}.

\subsection{Self-Supervised Loss (SSUL)}
Much of the success of BERT in NLP is replying on large scale self-supervised pretraining. Here, we propose to train the vision branch with a similar self-supervised loss. We first divide the input images into $N*N$ patches. Each patch is randomly masked by a ratio $p$. After that, the self-supervised loss predicts the mean color of the masked regions. To precisely predict the mean color of masked regions, the model will need to learn to extract useful information from contexts. Note that the self-supervised loss does not require any extra image annotations, hence can be easily scaled to arbitrary size of vision language training images. Formally, we define the SSUL as:

\begin{equation}
    SSUL =  \text{\textit{L2}}(Mean(\mathbf{m})| \mathbf{w}_{\backslash \mathbf{m}},M),
\end{equation}
where $\mathbf{m}$ is the masked areas, $\mathbf{w}_{\backslash \mathbf{m}}$ represents all other areas except for the masked area and $M$ represents the groundtruth mean color value of the masked area.

\subsection{Semantic Segmentation Loss (SEGL)}
One of the biggest challenges in vision language pretraining is that languages and images are features of different levels. Languages usually contain high-level semantic information. For example, one language word like ``car'' means one car object. In contrast, image information is low-level raw information. One image pixel does not have a specific meaning. Objects and semantic information are represented by groups of pixels. 

When the vision language models are trained with region feature input, the input vision features are already high-level semantic features, such that the vision and language correspondence can be learned fast. When the vision language models are trained with raw pixel input, the vision branch will be in charge of learning visual concepts.

To this end, we propose to help learning visual branch with an auxiliary Semantic Segmentation Loss (SEGL). Specifically, given vision output $X\in R^{\hat{H} \times \hat{W}}$, and groundtruth annotation $Y\in R^{H\times W}$, where $W$ and $H$ are image weight and height and $\hat{W}$ and $\hat{H}$ are feature map width and height, the SEGL is calculated by downsampling the annotation and calculating the cross-entropy loss between groundtruth and generated feature map:

\begin{equation}
    SEGL = \text{\textit{Cross-entropy}}(X,downsample(Y)).
\end{equation}

The SEGL is the same as the commonly used semantic segmentation loss. In normal semantic segmentation task, the networks usually up-sample the feature map and use semantic segmentation loss on the high-resolution feature map, which may include additional learnable parameters during up-sampling. In our task, we use SEGL as an auxiliary training target. Instead of up-sampling the feature map, we down-sample the groundtruth label map. The detailed network structure of Semantic Segmentation Loss is shown in Fig~\ref{fig:compare_methods} C.

\subsection{Knowledge Distillation Based Set Prediction Loss (SPL)}
Although the Semantic Segmentation Loss (SEGL) could provide semantic guidance to the network, it requires additional human annotation data, which is not preferable when training scales up. To overcome this problem, we propose the Knowledge Distillation Based Set Prediction Loss (SPL). SPL first runs a pretrained object detector to get $N$ object labels $y=\left\{y_i\right\}^N_{i=1}$ from the raw images. Unlike region feature based methods, the object detector is only run once before the pretraining step. The feature extraction process is not required during inference. 

\begin{figure*}[t]
\begin{center}
\includegraphics[width=1\linewidth]{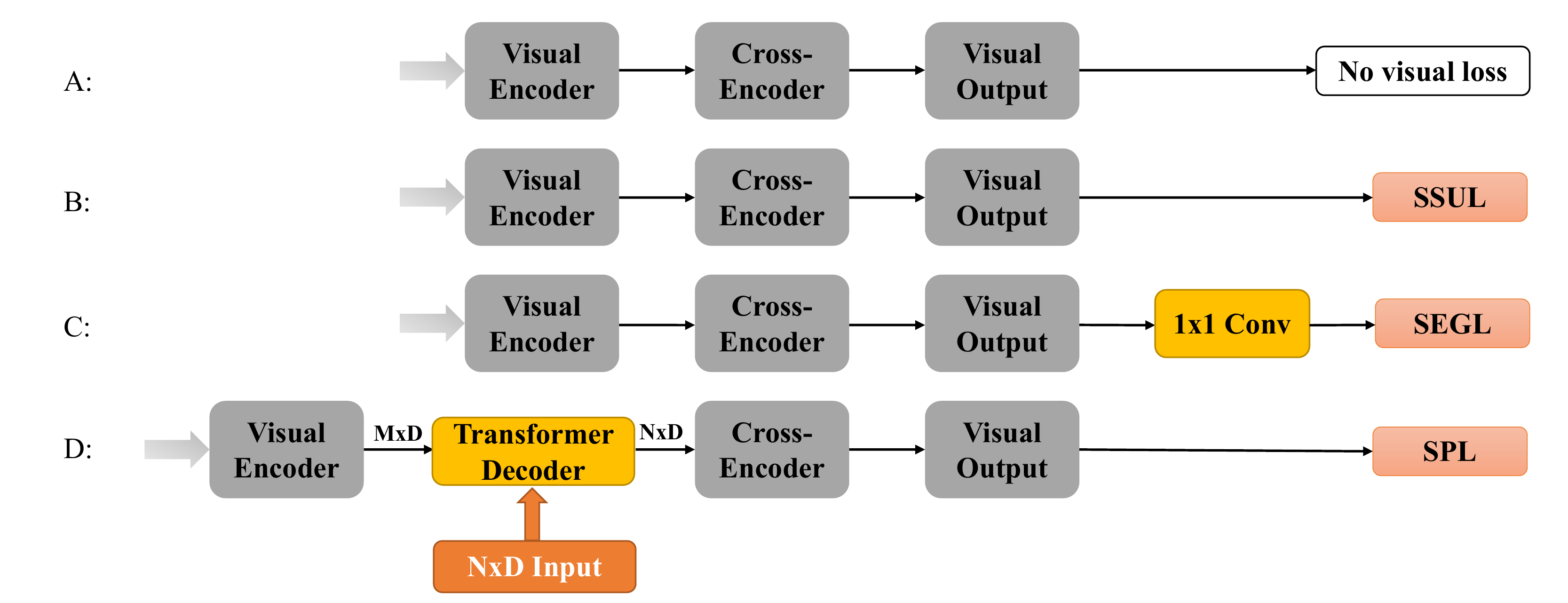}
	\caption{The detailed network structure for our proposed methods. A: No Visual Loss. B: Self-Supervised Loss. C: Semantic Segmentation Loss adds a one-layer convolution over the baseline method to map the feature to the category dimension. D: Set Prediction Loss. Given a large dimension visual feature, the set prediction loss takes a trainable N-dimensional input and uses a transformer decoder to map the feature dimension to N. Further cross encoder will be performed on the N-dimensional features. }
	\label{fig:compare_methods}
\end{center}
\end{figure*}

For the network part, given vision output $\hat{X}\in R^{\hat{N} \times \hat{M}}$, the network applies transformer decoder layers~\cite{devlin2018bert} to map the feature to $X\in R^{N}$, where $N<<\hat{N} \times \hat{M}$. The $N$ dimensional feature represents $N$ predicted objects. Then the network applies a Softmax layer and generates $N$ object predictions possibilities $\hat {p}=\left\{\hat {p}_i\right\}^N_{i=1}$. The detailed network structure is shown in Fig~\ref{fig:compare_methods} D.

SPL supervises the network training by matching the set of predictions $\hat{p}$ with the set of distillated labels $y$. The optimal bipartite matching problem is widely studied in multi-object tracking~\cite{huang2008robust,sahbani2016kalman} and is recently also extended to object detection~\cite{carion2020end}. The calculation of SPL is done in two steps: \textbf{find optimal matching between the two sets} and then \textbf{calculate object label cross-entropy loss}. 

\textbf{Optimal Matching.} The optimal matching between two sets is calculated by the Hungarian algorithm. Formally, given two sets $y=\left\{y_i\right\}^N_{i=1}$ and $\hat {p}=\left\{\hat {p}_i\right\}^N_{i=1}$, The Hungarian algorithm finds bipartite matching by minimizing the cost of assigning correct labels $-\hat{p}_{\sigma(i)}(c_i)$. Formally, the optimal matching $\hat{\sigma}$ is calculated by:
\begin{equation}
\label{eq:matching}
    \hat{\sigma} = \argmin_{\sigma} \sum_{i}^{N} [-\hat{p}_{\sigma(i)}(c_i)],
\end{equation}
where $c_i$ is the category of object index $i$ and $\sigma$ is a permutation of the prediction $\hat {p}$ and $\hat{\sigma}$ represents the optimal matching.

\textbf{Object Label Classification Loss.} After the optimal bipartite matching is found, we perturb the generated labels with calculated optimal bipartite matching $\hat{\sigma}$ and continue to calculate the SPL loss as an object label cross-entropy loss between the matched prediction and distillated labels:

\begin{equation}
\label{eq:matching_2}
    SPL = \text{\textit{Cross-entropy}}(\hat{p}_{\hat{\sigma}},y).
\end{equation}

To summarize and compare our proposed methods, we list extra computation cost, extra data, scalability, and inference speed of our methods vs baseline in Table ~\ref{table:compare_methods}.

\begin{table*}[]
\small 
\centering
\resizebox{0.9\textwidth}{!}{
\begin{tabular}{ccccc}
\hline

Method                & Extra Comp Cost      & Extra Data            & Scalability & Inference Speed \\ \hline
\hline
No Visual Loss        & No                          & No                    & Yes                                 & -               \\
SSUL       & No                          & No                    & Yes                                 & Same            \\
SEGL & 1 layer 1x1 Conv            & Segmentation Mask     & Partial                                  & Same            \\
SPL       & 1 layer transformer decoder & Pretrained detector & Yes                                 & Faster          \\ \hline \\
\end{tabular}

}

\caption{Comparison of the three proposed methods. We show that the self-supervised loss is the most general method that requires no extra computational cost and no extra training data, hence can be easily scaled to arbitrary training image size. Semantic segmentation loss requires only a one-layer convolution during pretraining. The set prediction loss uses an extra one-layer transformer decoder to reduce feature dimension and is faster during inference.}
\label{table:compare_methods}
\end{table*}

\begin{table*}[t]
\small 
\centering
\setlength{\tabcolsep}{7mm}{%
\resizebox{0.8\textwidth}{!}{
\begin{tabular}{ccccc}
\hline
\multicolumn{1}{c}{Method} &Inference Time (ms) & VQA(\%)  \\ \hline\hline
Baseline (ViT)  &40& 68.95        \\ 
ResNet &40& 68.52      \\
Baseline + SSUL &40& 70.26         \\ 
Baseline + SEGL &40& \textbf{70.92}      \\ 
ResNet + SEGL &40& 70.53        \\
Baseline + SPL(36) &34&69.41    \\ 
Baseline + SPL(100) &37& 70.20       \\ 

\hline \\
\end{tabular}%
}
}
\caption{Ablation experiments of our proposed methods on VQA tasks. We do experiments on VQAv2. Our proposed method could improve performance on both backbones.}
\label{table:ablation}
\end{table*}

\begin{table*}[t]
\small 
\centering
\resizebox{1\textwidth}{!}{%
\begin{tabular}{cccccccc}
\hline
\multicolumn{1}{c}{Method} &Inference Time (ms) &  TR (COCO 5k)  &  IR (COCO 5k) & TR (COCO 1k) &  IR (COCO 1k)\\ \hline\hline
Baseline (ViT)   &40& 0.25  & 0.12  & 0.40& 0.33     \\ 
ResNet &40& 0.27  & 0.11  & 0.38& 0.32     \\
Baseline + SSUL &40& 0.48 & 0.33  & 0.62& 0.51         \\ 
Baseline + SEGL &40& \textbf{0.50}& \textbf{0.37}  & \textbf{0.67}& \textbf{0.54}      \\ 
ResNet + SEGL &40& \textbf{0.50} & \textbf{0.37} & 0.65& 0.535     \\
Baseline + SPL(36) &34&0.42& 0.25 &0.55& 0.44    \\ 
Baseline + SPL(100) &37& 0.42& 0.26 &0.61& 0.495       \\ 

\hline \\
\end{tabular}%
}
\caption{Ablation experiments of our proposed methods on retrieval tasks. We report the accuracy at R@1. We do experiments on zero-shot text retrieval and zero-shot image retrieval. We show that all of our proposed methods greatly improve the performance on down-stream tasks. Among all of them, the semantic segmentation loss gives the best performance. We also show that the training difficulty of pixel input models exists for all types of visual backbones (ResNet and ViT).}
\label{table:ablation2}
\end{table*}

\begin{table*}[t]
\begin{center}
\resizebox{1\textwidth}{!}{
\begin{tabular}{l c c c c c c c c c c c c  c c}
  \hline

      \multirow{2}*{Models} & \multirow{2}*{Images} & \multirow{1}*{Pretrain} &\multirow{1}*{Inference}    & \multicolumn{2}{c}{VQAv2}   & \multicolumn{2}{c}{NLVR2} &\multicolumn{3}{c}{VCR}\\ 
      \cmidrule(r){5-6} \cmidrule(r){7-8} \cmidrule(r){9-11} 
      &  &GPU hrs &Time (ms)& test-dev & test-std  & val  & test-P & Q $\rightarrow$ A& QA $\rightarrow$ R& Q $\rightarrow$ AR\\
      \hline
          ViLBERT~\cite{lu2019vilbert}  & 3m & -&$\sim$900 & 70.55  &  70.92   & -& - & -& -&- \\ 

    LXMERT~\cite{tan2019lxmert} & 180k &$\sim$800 & $\sim$900& 72.42  &  72.5   &74.9& 74.5 & -& -&- \\ 

    UNITER-base~\cite{chen2019uniter} & 4.2m  &882 & $\sim$900& \textbf{72.70}  &  \textbf{72.91}   &\textbf{75.85}&75.80& 74.56&77.03&57.76   \\
   Pixel-BERT (R50)~\cite{huang2020pixel} & 200k  &  $\sim$4000 & $\sim$60& 71.35  & 71.42  &71.7& 72.4& -& -&-  \\
   VilT~\cite{kim2021vilt} & 4.1m  &  $\sim$4600 & $\sim$15& 71.26 & - &75.70& \textbf{76.13} & -& -&- \\
    Ours (SEGL) & 230k&  \textbf{470}&$\sim$40 & \textbf{71.69} & \textbf{72.03}   &72.3& 72.9 & \textbf{75.37}   &\textbf{77.86}& \textbf{58.33}\\ 
    \hline
           
\end{tabular}
}

\end{center}
\caption{Comparison with other vision-language pre-training models on VQAv2, NLVR2, and VCR. The pretrain GPU hours of Pixel-BERT are approximated based on pretrain image size and epochs. All pretrain GPU hours are on V100 GPUs. Inference time is based on P40 GPUs. We show that our method could achieve competitive performance on the two datasets and run more than 10x faster compared with other region feature models. Compared with raw pixel input models, our method trains 10 times faster and achieves better accuracy.}
\label{table:tab_vlp}
\end{table*} 

\begin{table*}[]
\resizebox{1\textwidth}{!}{
\begin{tabular}{llllllllllllll}
\hline
\multicolumn{1}{c}{\multirow{3}{*}{Model}}                   & \multicolumn{1}{c}{\multirow{3}{*}{Images}} & \multicolumn{6}{c}{Flickr 30k (1k image)}                                                                                                                   & \multicolumn{6}{c}{MSCOCO (1k image)}                                                                                                                       \\ \cline{3-14} 
\multicolumn{1}{c}{}                                         & \multicolumn{1}{c}{}                        & \multicolumn{3}{c}{Image Retrieval}                                          & \multicolumn{3}{c}{Text Retrieval}                                           & \multicolumn{3}{c}{Image Retrieval}                                          & \multicolumn{3}{c}{Text Retrieval}                                           \\
\multicolumn{1}{c}{}                                         & \multicolumn{1}{c}{}                        & \multicolumn{1}{c}{R@1} & \multicolumn{1}{c}{R@5} & \multicolumn{1}{c}{R@10} & \multicolumn{1}{c}{R@1} & \multicolumn{1}{c}{R@5} & \multicolumn{1}{c}{R@10} & \multicolumn{1}{c}{R@1} & \multicolumn{1}{c}{R@5} & \multicolumn{1}{c}{R@10} & \multicolumn{1}{c}{R@1} & \multicolumn{1}{c}{R@5} & \multicolumn{1}{c}{R@10} \\ \hline
ViLBERT~\cite{lu2019vilbert}           &  3m                                           &           31.83              &     61.12                    &     72.80                     &           -              &       -                  &      -                    &-                         &           -              &          -                &    -                     &           -              &           -               \\
Unicoder-VL~\cite{li2020unicoder}    &   3.8m                                          &        48.40                 &             76.00            &      85.20                    &     64.30                    &    85.80                     &    92.30                      &          -               &        -                 &        -                  &               -          &         -                &        -                  \\
ImageBERT ~\cite{qi2020imagebert}    &    10m                                         &       54.3                  &             79.6            &                87.5          &               70.7          &      90.2                   &           94.0               &            53.6             &     83.2                    &             91.7             &         67.0                &          90.3               &    96.1                      \\
UNITER-base~\cite{chen2019uniter}   &      4.2m                                       &     \textbf{66.16}                    &         \textbf{88.40}                &                \textbf{92.94}          &        80.70                 &      95.70                   &              \textbf{98.00}            &          -               &          -               &      -                    &             -            &            -             &           -               \\

                                   Ours (SEGL)                        &     \textbf{230k}                                        &                   62.46          &              86.71         &                  92.80       &                     \textbf{82.70}    &                 \textbf{96.50}         &                    \textbf{98.00}      &                  \textbf{54}    & 
 \textbf{86}
                      &    \textbf{94.8}
                      &    \textbf{67.1}
                     &        \textbf{92.2}
               & \textbf{97.1}
 \\
 \cline{1-14} 
\end{tabular}
}
\\
\caption{Experiments of Zero-shot Image and Language Retrieval on Flicker 30k dataset and COCO 1k dataset. We achieve competitive results on Flicker and COCO datasets with far fewer training images.}
\label{table:tab_vlp_2}
\end{table*}

\begin{table*}[]
\centering
\setlength{\tabcolsep}{2.3mm}{
\resizebox{0.8\textwidth}{!}{
\begin{tabular}{llllllll}
\hline
\multicolumn{1}{c}{\multirow{3}{*}{Model}}                   & \multicolumn{1}{c}{\multirow{3}{*}{Images}}                                                                                                               & \multicolumn{6}{c}{MSCOCO (5k image)}                                                                                                                       \\ \cline{3-8} 
\multicolumn{1}{c}{}                                         & \multicolumn{1}{c}{}       & \multicolumn{3}{c}{Image Retrieval}                                          & \multicolumn{3}{c}{Text Retrieval}                                           \\
\multicolumn{1}{c}{}                                         & \multicolumn{1}{c}{}             & \multicolumn{1}{c}{R@1} & \multicolumn{1}{c}{R@5} & \multicolumn{1}{c}{R@10} & \multicolumn{1}{c}{R@1} & \multicolumn{1}{c}{R@5} & \multicolumn{1}{c}{R@10} \\ \hline
ViLBERT~\cite{lu2019vilbert}           &  3m                                                &31.86                        &        61.12              &         72.80              &    -                    &          -              &          -               \\
ImageBERT ~\cite{qi2020imagebert}    &    10m                                           &           32.3             &    59.0                  &            70.2            &         44.0                &          71.2               &   80.4                     \\

VilT ~\cite{kim2021vilt}    &    4.1m                                           &           \textbf{40.4}             &   \textbf{70.0}                 &            \textbf{81.1}            &        \textbf{56.5}                &          \textbf{82.6}               &   \textbf{89.6}                     \\

                                   Ours (SEGL)                        &     \textbf{230k}                                          &                  37.2    & 
 63.3
                      &    74.4
                      &    49.8
                     &        75.9
               & 83.1
 \\
 \cline{1-8} 
 \\
\end{tabular}
}
}
\\

\caption{Additional Experiments on Zero-shot Image and Language Retrieval on COCO 5k image. We still achieve competitive results with fewer training images.}
\label{table:tab_vlp_3}
\end{table*}

\begin{figure*}[]
\begin{center}
\includegraphics[trim=0 150 0 100,width=1\linewidth]{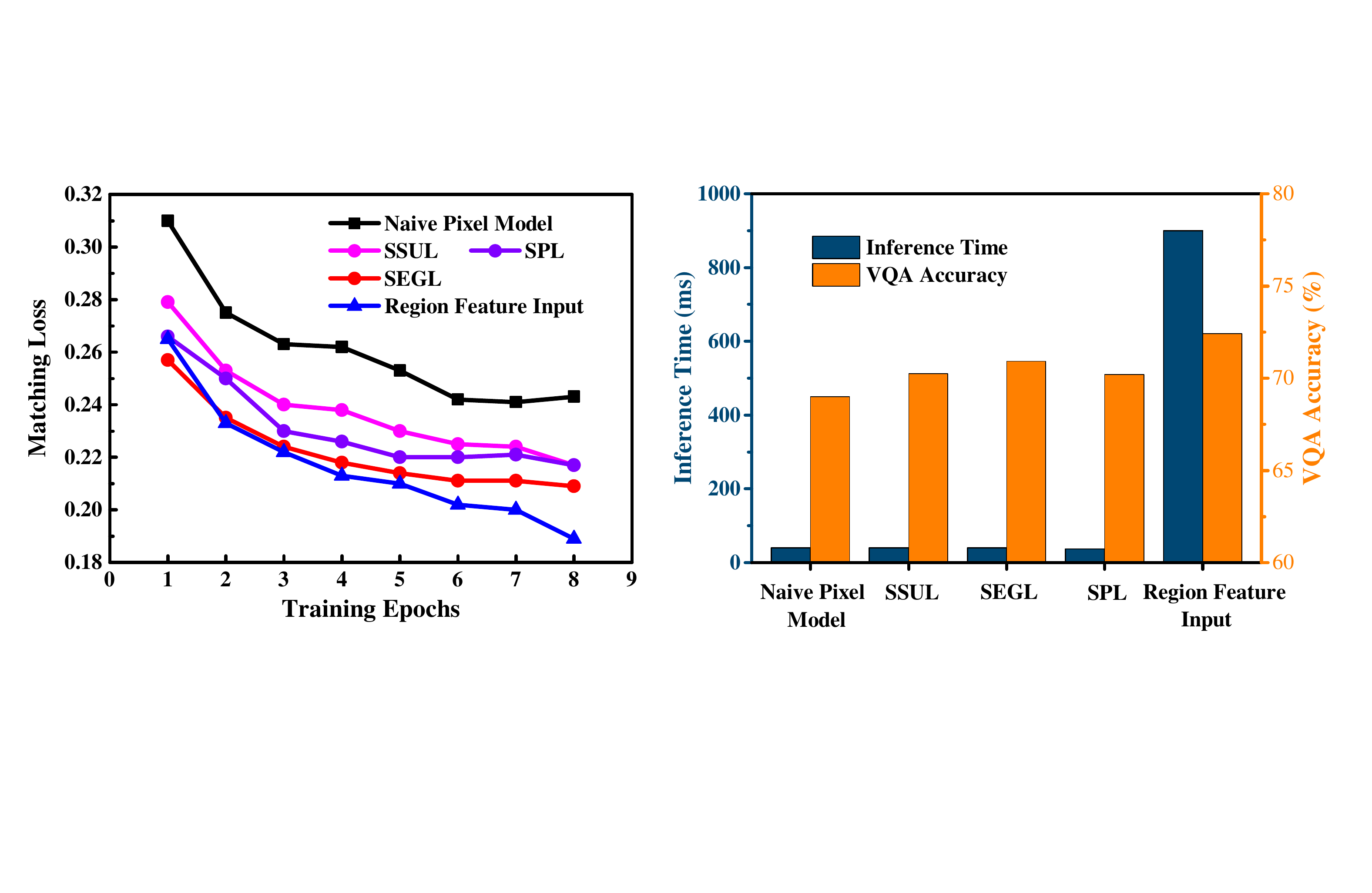}
	\caption{Left: The convergence speed (reducing speed of matching loss) on validation set. Right: Inference time and VQA Accuracy. We show that when applying all our three losses on the baseline model, we could witness a great boost up in convergence speed. Among all of them, the SEGL gives the best convergence speed, which is almost the same as a region feature model. As for inference speed, our models achieve the same or faster inference speed as a naive baseline model and more than 10 times faster than a region feature model. Accuracy-wise, our method achieves similar performance as the region feature model and is better than the naive baseline.}
	\label{fig:loss_all}
\end{center}
\end{figure*}

\begin{figure*}[t]
\begin{center}
\resizebox{1.05\textwidth}{!}{
\includegraphics{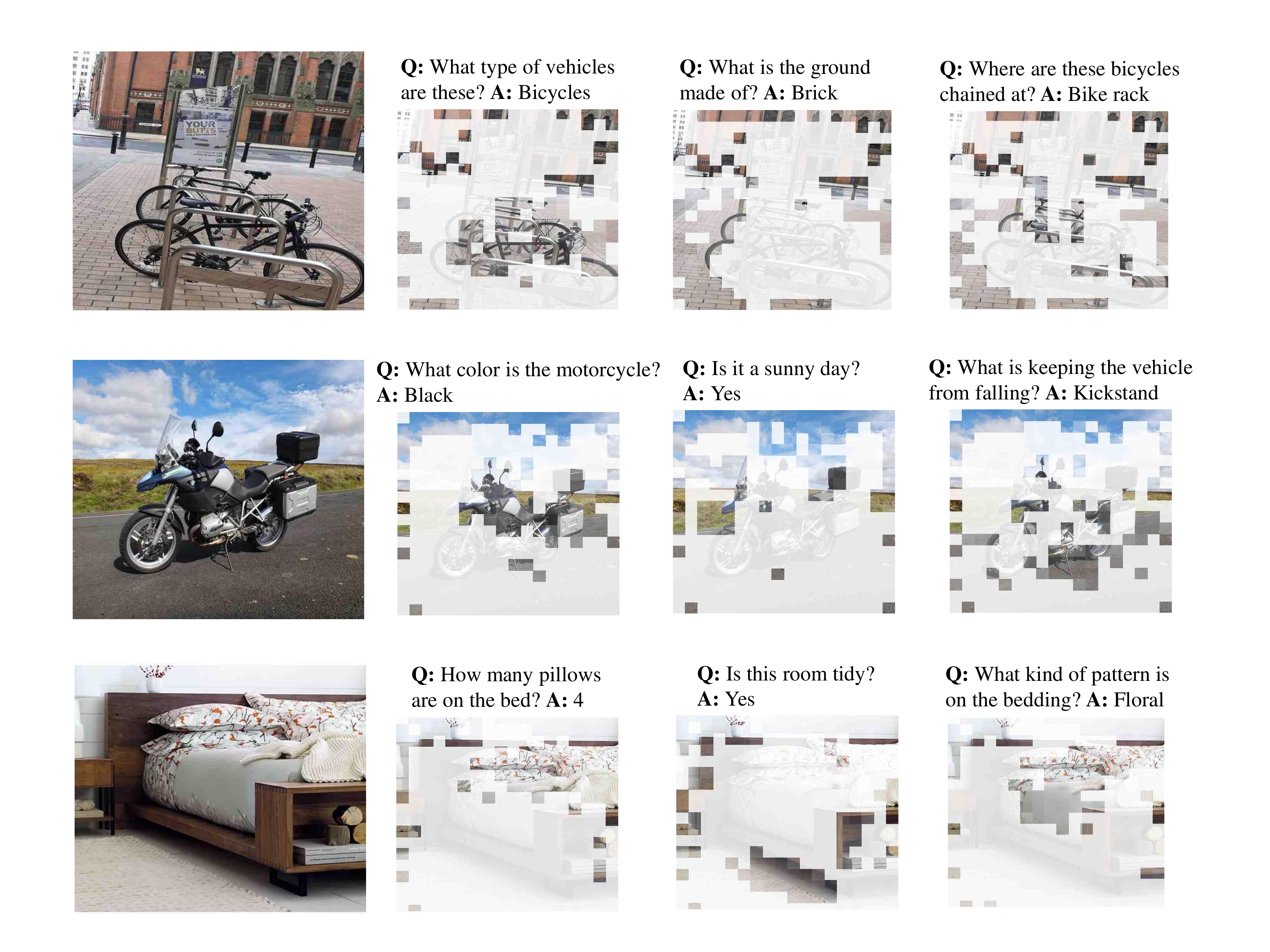}
}
	\caption{Visualization of attention maps. First row: the end-to-end model could easily identify no-object items, for example, the ground. Second row: the end-to-end model could attend to small objects, for example, the kickstand. Third row: The grid feature model is weak at counting problems.}
	\label{fig:7}
\end{center}
\end{figure*}

\section{Experiment}

\subsection{Implementation Details}
\subsubsection{Network Structure} 

We follow the network structure of 2-stream models ViLBERT~\cite{lu2019vilbert} and LXMERT~\cite{tan2019lxmert}, remove the visual feature encoder branch, and replace it with a grid feature extractor. To prove that the training difficulty is consistent for all types of visual backbones. We do experiments for both CNN model ResNet and Vision Transformer model. For CNN model, we use ResNet50~\cite{he2016deep}. For Vision Transformer model~\cite{dosovitskiy2020image}, we use a nine layers transformer model with patch size 32. For both of the backbones, we resize the input image size to 384x384, which is only 40\% of the image size compared with Pixel-BERT~\cite{huang2020pixel}. For language input, we follow the original BERT~\cite{devlin2018bert} to tokenize the sentence into tokens with wordpiece tokenizer and append a [CLS] token and a [SEP] token before and after the sentence. The Cross-Modality Encode contains 5 layers of cross-modality layers. We load pretrained weights for language branch and vision backbones.

\subsubsection{Training Data} 

We only use image-sentence pairs in COCO~\cite{chen2015microsoft}, Flickr~\cite{flickrentitiesijcv} and Visual Genome~\cite{krishna2017visual}, which contains 230k images and 5 million image sentence pairs. We compare our training data with other models in Table~\ref{table:tab_vlp}. We strictly follow the data splitting process  in UNITER~\cite{chen2019uniter} to make sure that there’s no overlap between the training set and testing set. For Semantic Segmentation Loss (SEGL), we use the annotations provided in COCO Stuff~\cite{caesar2018cvpr}. The semantic segmentation groundtruth contains 172 classes of pixel labels on 100k of our training images, which covers around 50\% of pretraining image data and only 10\% of pretraining image-sentence pairs. For Knowledge  Distillation  Based  Set  Prediction Loss (SPL), we use the faster-rcnn model provided in Bottom-Up attention~\cite{anderson2018bottom} to extract object labels. 
\subsubsection{Data Augmentation} 
Pixel input models also enable online data augmentation during pretraining. We propose a novel \textbf{Description aware data augmentation method}. Horizontal flip can not be applied to all images, especially those images with compositional descriptions. To solve this problem, we build an empirical dictionary with compositional keywords (e.g. left, right, etc.). If a description does not contain keywords in the dictionary, we perform horizontal flip with a chance of 50\%. The data augmentation is only used in pretraining. No data augmentation is used during finetuning.  \textbf{Random Resize and Crop.} We randomly resize the input image between 0.8-1.2 and crop a 384x384 patch in the resized image. 

\subsubsection{Implementation Details of SSUL, SEGL and SPL} 

\textbf{1) SSUL:} For self-supervised loss, we divide the input image size into 32x32 patches. Each patch is randomly masked at a ratio of $p=0.15$, which is the same as language words are masked. \textbf{2) SEGL:} For semantic segmentation loss, to save GPU memory, we resize the groundtruth annotations to the same dimension as the feature map and not vice versa. For vision transformer, we resize the groundtruth annotations to 12x12. For ResNet 50, we resize the feature map to 24x24. \textbf{3) SPL:} We try to extract both 36 objects and 100 objects. Note that a typical vision transformer model produces a feature dimension of 144, so even the 100 object version requires less memory than the naive baseline models. For cross-attention modules, the network takes 36 or 100 object features as input. This design also makes sure the memory usage of cross-attention modules is consistent for whatever image input size. 

For all of the proposed methods and baseline models, we pretrain the models for a maximum of 15 epochs and choose the checkpoint with the lowest validation loss for finetuning.

\subsection{Finetuning Setting}
In this section, we describe the finetuning settings on downstream tasks.

\subsubsection{VQAv2}
The task of Visual Question Answering is to answer a given question based on the content of an image. The VQA~\cite{goyal2017making} task is usually formatted as a classification problem on all possible answers. During finetuning, we add a two-layer MLP on top of the language features of the [CLS] position and finetune the MLP together with the backbone models.

\subsubsection{NLVR2}
The Natural Language for Visual Reasoning for Real dataset~\cite{suhr2017corpus} is designed to examine the reasoning ability of the networks. Specifically, given two images, the network will be asked to tell the statement is true or false by comparing the two images. To finetune the NLVR2 task, we use the easiest solution that processes image-language pairs separately and concatenates the [CLS] features. After concatenation, we use a binary classifier to classify if the description is true. 

\subsubsection{VCR}

Visual Commonsense Reasoning (VCR)~\cite{zellers2019recognition} is a dataset that asks commonsense multiple-choice questions based on an image. There are two sub-tasks: the question-answering task (Q $\rightarrow$ A) and the justification task (QA $\rightarrow$ R). The combined task is represented as Q $\rightarrow$ AR. We follow the same finetuning protocol as UNITER~\cite{chen2019uniter}. We use each question-answer pair as the input to the language branch and concatenate the [CLS] features of the pairs. A classifier is used to select the best answer based on the concatenated features.

\subsubsection{Zero-Shot Image and Text Retrieval}
The zero-shot image (text) retrieval task is to find the best matched image (text) in the database given a text (image) without extra finetuning. The retrieval technique is commonly used in modern image-language search engines. To do this, we run inference using pretrained weights on all image and language pairs in the database and rank their matching scores. The experiments are carried out on both Flickr 30k~\cite{flickrentitiesijcv} and COCO dataset~\cite{chen2015microsoft}.

\subsection{Ablation Study}

We start by building a naive version of pixel input vision language pretraining model using vision transformer~\cite{dosovitskiy2020image} backbone. Same as Pixel-BERT~\cite{huang2020pixel}, the naive version uses only a matching loss and the language modeling loss. We compare our proposed three losses with the baseline version on three tasks. A detailed ablation study is shown in Table~\ref{table:ablation} and Table~\ref{table:ablation2}. For all ablation experiments, we use finetuning input size 600. Following~\cite{kim2021vilt}, the inference time is calculated by the running speed on Nvidia P40 GPU.

We show that all of our proposed methods surpass the baseline model with a large margin. Within the proposed methods, the Semantic Segmentation Loss achieves the best performance in general. It proves that additional human annotations do help to improve pretraining vision language models. Self-supervised loss and set prediction loss also produce good results on down-stream tasks -- less than 1 percent lower than semantic segmentation loss on VQA. Among all of them, the set prediction loss has a relatively faster process speed and lower memory usage. For the cross-modality module, the set prediction loss uses 36 or 100 visual features, while other vision transformer models use 324 (18*18) visual features.

By comparing ResNet~\cite{he2016deep} models and vision transformer models~\cite{dosovitskiy2020image}, we conclude that the pretraining difficulty exists regardless of the backbone choice and the backbones won't affect the performance of our proposed losses.

\subsection{Comparison with Other Methods}
We compare our best result (Vision Transformer + SEGL) with region feature models and pixel input models in Table~\ref{table:tab_vlp}, Table~\ref{table:tab_vlp_2} and Table~\ref{table:tab_vlp_3}. For all experiments in this section, we continue to use image size 600x600 except for VQA. For VQA we use image size 800x800 to further improve performance, same as Pixel-BERT~\cite{huang2020pixel}.

Compared with region feature models, when trained with fewer GPU hours, our method could achieve almost the same performance on various datasets. During inference, an end-to-end model could run more than 10x faster than any region feature based model. 

Compared with other end-to-end models, our model is trained for only 1/10 of the pretraining GPU hours and gets competitive or even better finetuning performance.

\subsection{Visualization}

\subsubsection{Visualization of Convergence Speed and Finetuning Accuracy}
We plot the pretraining matching loss and finetuning performance in Fig~\ref{fig:loss_all}. We show that our proposed methods greatly improve convergence speed during pretraining. Among all of them, the semantic segmentation loss converges at the fastest speed. 

\subsubsection{Visualization of Attention Map}

To further prove an end-to-end model could learn coherent vision language representations, we plot the attention map of the last layer of cross-attention modules in Fig~\ref{fig:7}. The attention map shown is the average value of all attention heads. Specifically, we use the finetuned model on VQAv2 dataset as an example. We show the attention map of one image given different questions. 

First, we show that the pixel input model could attend to both objects and no-object items, for example, the ground. Answering no-object questions is difficult for a region feature based model. 

In the second example, we show an example that the pixel input model could attend to small items like the kickstand. Detecting such small items is challenging for an object detector, therefore it's difficult for the region feature based models to answer questions related to small items. 

Third, grid features by default don't understand numbers. When asked about counting problems, although our end-to-end model could learn to attend to the right area, it does not give a correct answer (3).

\section{Conclusion and Future Works}
In this paper, we have targeted the difficulty in end-to-end vision language pretraining i.g. slow convergence speed and low finetuning accuracy. We propose three losses. All of them could surpass the baseline naive model with a large margin. We  show  that  our  proposed  methods could  achieve  similar  performance  when  trained  with  the same GPU hours as region feature models.  During inference time, our model runs 10 times faster than the region feature models.  Compared with other pixel input vision language pretraining models, our models only require 1/10 of the pre-training GPU hours.  We also evaluate different vision back-bones,  e.g.   the  CNN  backbones  and  Vision  Transformer backbones.

\textbf{Future works} can be done by increasing the scale of pretraining. Our proposed self-supervised loss and set-prediction loss enable pain-free scalability. By using more training images from Conceptual Caption~\cite{sharma2018conceptual} and other datasets, a pixel input end-to-end vision language model could be further improved. 

More works could also be done by exploring more complicated data augmentations. As mentioned earlier in the paper, an end-to-end vision language model enables online data augmentation. In this paper, we only use simple augmentation methods like random cropping and flip in pretraining. No data augmentation is used in finetuning. Methods like Rand-augment~\cite{cubuk2020randaugment} could be explored to further improve performance. 

Jointly using the proposed losses could be further explored. Currently, we haven’t witnessed improvements by jointly using more than one proposed losses. Future works can be done by exploring the proper weights of the proposed methods. 

\section*{Acknowledgments}
This research is supported by the National Research Foundation, Singapore under its AI Singapore Programme (AISG Award No: AISG-RP-2018-003), the MOE AcRF Tier-1 research grants: RG95/20, and the OPPO research grant. This research is also supported by the Agency for Science, Technology and Research (A*STAR) under its AME Programmatic Funds (Grant No. A20H6b0151).

%
%
\bibliographystyle{IEEEtran}
\bibliography{egbib}

\begin{IEEEbiography}[{\includegraphics[width=1in,height=1.25in,clip,keepaspectratio]{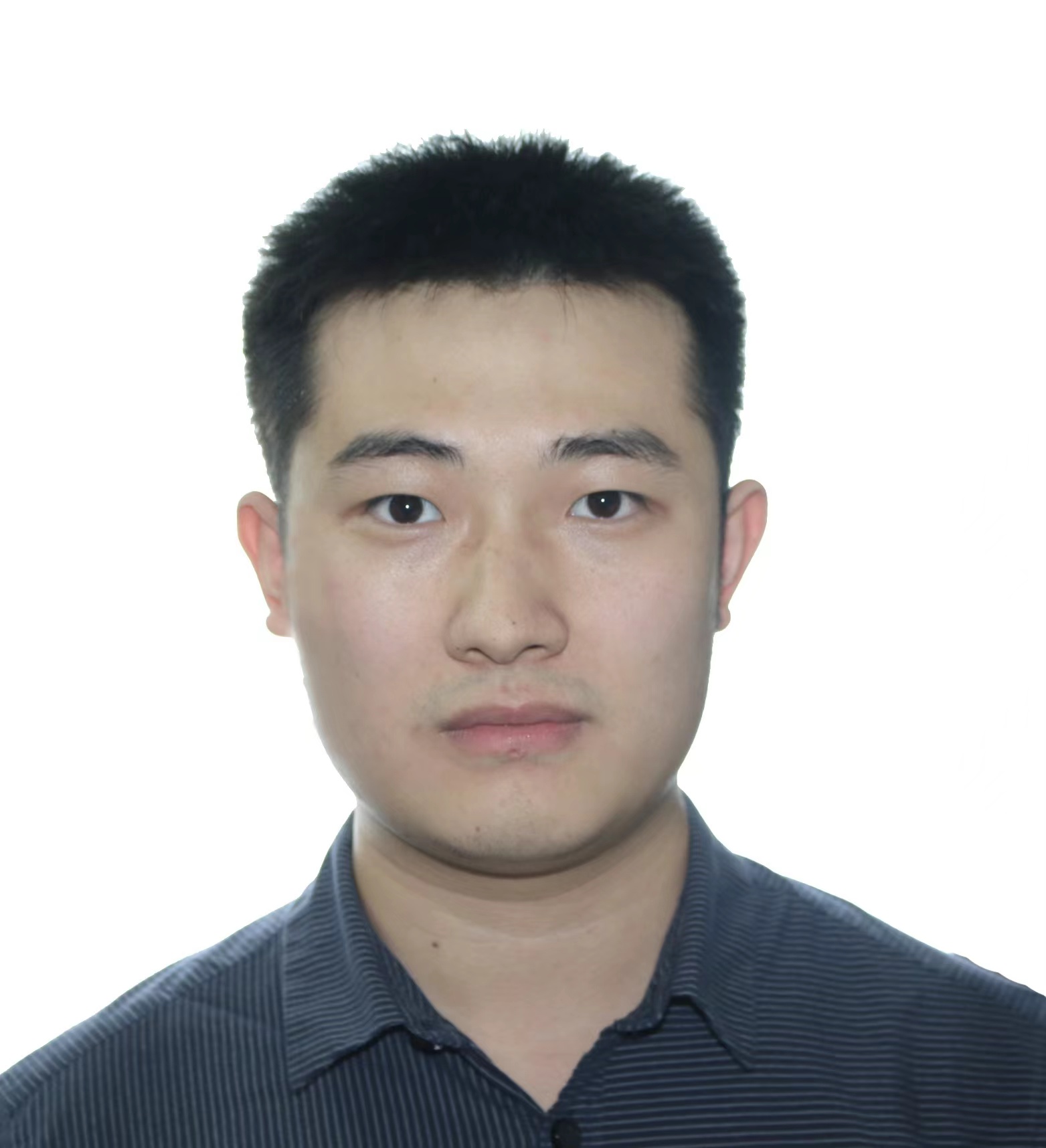}}]{Xiaofeng Yang} is a PhD student at the School of Computer Science and Engineering, Nanyang Technological University, Singapore. His research interests are in computer vision and machine learning.
\end{IEEEbiography}

\begin{IEEEbiography}[{\includegraphics[width=1in,height=1.25in,clip,keepaspectratio]{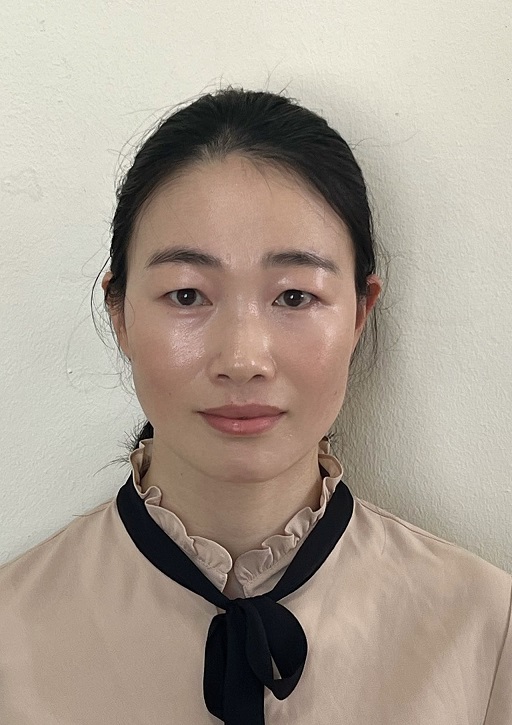}}]{Fayao Liu} is a research scientist at Institute for Infocomm Research (I2R), A*STAR, Singapore. She received her PhD in computer science from the University of Adelaide, Australia in Dec. 2015. Before that, she obtained her B.Eng. and M.Eng. degrees from National University of Defense Technology, China in 2008 and 2010 respectively. She mainly works on machine learning and computer vision problems, with particular interests in self-supervised learning, few-shot learning and generative models. She is serving as an associate editor for IEEE Transactions on Circuits and Systems for Video Technology (TCSVT).
\end{IEEEbiography}

\begin{IEEEbiography}[{\includegraphics[width=1in,height=1.25in,clip,keepaspectratio]{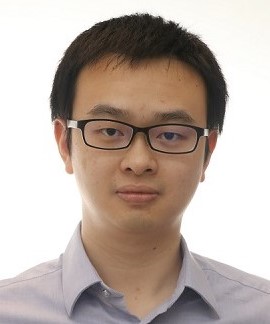}}]{Guosheng Lin} is an Assistant Professor at the School of Computer Science and Engineering, Nanyang Technological University, Singapore. He received his PhD degree from The University of Adelaide in 2014. His research interests are in computer vision and machine learning.
\end{IEEEbiography}

\end{document}